# Consistency of AI-Generated Exercise Prescriptions: A Repeated Generation Study Using a Large Language Model

Kihyuk Lee[*]

Data Convergence Team, Office of Hospital Information, Seoul National University Bundang Hospital, Seongnam, Republic of Korea

**Correspondence to**
Kihyuk Lee, PhD
Data Convergence Team, Office of Hospital Information, Seoul National University Bundang Hospital, 172 Dolma-ro, Bundang-gu, Seongnam, 13605, Republic of Korea

**Abstract**

**Background:** Large language models (LLMs) have recently been explored as tools for generating personalized exercise prescriptions. While previous studies have demonstrated their feasibility and safety, the consistency of outputs under identical conditions remains insufficiently examined.

**Objective:** This study aimed to evaluate the intra-model consistency of LLM-generated exercise prescriptions using a repeated generation design.

**Methods:** A total of six clinical scenarios, including both clinical patient groups and healthy individuals, were used to generate exercise prescriptions using Gemini 2.5 Flash. For each scenario, 20 outputs were generated under identical prompt conditions (total n = 120). Consistency was assessed across three dimensions: (1) semantic consistency using SBERT-based cosine similarity, (2) structural consistency based on the FITT (Frequency, Intensity, Time, Type) principle using an AI-as-a-judge approach, and (3) safety expression consistency, including both inclusion rates and sentence-level quantification.

**Results:** Semantic similarity was consistently high across scenarios (mean cosine similarity: 0.879–0.939), indicating stable textual outputs. Higher consistency was observed in clinically constrained cases compared to high-function scenarios. Structural analysis revealed that frequency showed consistent patterns, whereas variability was observed in quantitative components, particularly exercise intensity. In resistance training prescriptions, unclassifiable intensity expressions were observed in 10–25% of outputs. Safety-related expressions were included in 100% of outputs; however, the number of safety-related sentences varied significantly across scenarios (H=86.18, p<0.001), with clinical cases generating substantially more safety expressions than healthy adult cases.

**Conclusions:** While LLM-generated exercise prescriptions demonstrated high semantic consistency and generally aligned with guideline-based structures, variability in key quantitative components revealed limitations in generating stable numerical prescriptions. These findings indicate that the reliability of LLM outputs depends substantially on prompt structure, and that additional structural constraints and expert validation are needed before clinical deployment.

## 1. Introduction

The advancement of large language models (LLMs) has opened new possibilities in the field of exercise prescription. LLMs are capable of generating structured exercise recommendations that account for individual health status, disease characteristics, and contraindications, and their potential as decision-support tools in settings where access to specialists is limited has been widely discussed [1]. A growing body of expert evaluation studies has examined LLM-generated prescriptions across various contexts, including resistance training programs, running training plans, and exercise prescriptions for patients with diabetes [2-4], and a recent scoping review confirmed that research in this area is transitioning from an exploratory phase to one of systematic validation [5].

Exercise prescription is fundamentally distinct from general health information provision, and in high-risk clinical contexts involving multiple chronic conditions, functional limitations, and contraindications, verification of not only output quality but also reproducibility is essential. Given these characteristics, exercise prescription in patients with comorbidity requires complex clinical reasoning that may lead to variability in professional judgment [6], and the application of AI-based systems demands even more rigorous validation of reproducibility. In this study, reproducibility refers to the capacity of an LLM to produce stable outputs under identical conditions, and consistency refers to the degree of similarity across repeated outputs as its measurable proxy. Output variability in this context carries direct clinical implications: the same patient profile may yield structurally different prescriptions across repeated generations, raising concerns about decision-making consistency in real-world deployment. Choi et al. [7] evaluated LLM-generated exercise prescriptions under a structured prompt framework and reported that prompt structuring improved safety and guideline adherence, but noted low inter-rater agreement (ICC (2,3) = 0.139) and difficulty securing statistical independence among repeated outputs, proposing systematic consistency analysis as a future direction.

LLMs may produce variable outputs across repeated generations under identical prompts due to their probabilistic token generation mechanism [8], representing a direct concern for clinical reliability in exercise prescription. The LLM-as-a-Judge paradigm, in which a separate LLM evaluates outputs according to a predefined rubric, has been proposed as a scalable and reproducible complementary assessment approach [9,10]. However, limitations such as self-preference bias and length bias have been reported [11], and more stable application is expected in domains where clear binary criteria exist, such as adherence to the FITT principle and inclusion of safety-related content.

Building on the structured prompt framework developed by Choi et al. [7], this study aims to quantify the repeated generation consistency of LLM-generated exercise prescriptions across three dimensions: semantic similarity (SBERT-based cosine similarity), structural consistency based on the FITT principle, and safety expression consistency. By introducing reproducibility and consistency as evaluation axes that extend beyond accuracy and safety-focused assessments, this study provides a foundational basis for ensuring the reliability of LLM-based exercise prescription systems and for developing automated evaluation methodologies.

## 2. Materials and Methods

*2.1. Study Design and Overview*

This study was conducted to evaluate the intra-model consistency of exercise prescriptions generated by a large language model (LLM) under controlled conditions. A repeated generation design was applied, in which identical clinical scenarios and prompts were repeatedly submitted to produce multiple outputs, allowing for the assessment of within-model variability.

*2.2. Clinical Scenario Development*

Six clinical scenarios were used: three adopted from a previous study [7] representing high-risk conditions across metabolic, musculoskeletal, and oncological domains, and three newly developed to extend clinical variability by including a multimorbidity case with cardiovascular risk factors and two healthy adult cases targeting fat loss and muscle hypertrophy, respectively. All scenarios were hypothetical, constructed without real patient information, and designed to reflect clinically relevant factors including age, sex, disease characteristics, functional limitations, and exercise goals. Key characteristics are presented in Table 1, and detailed clinical information is provided in Supplementary Material 1.

**Table 1.** Summary of Clinical Scenarios Used for Repeated Exercise Prescription Generation

| Clinical Cases | Sex | Age (Years) | Primary Condition(s) | Key Functional Limitation or Risk | Baseline Physical Activity Level | Primary Exercise Goal |
|---|---|---|---|---|---|---|
| Case 1 (T2DM + Obesity) | Male | 55 | Type 2 diabetes mellitus, obesity | Limited exercise experience, mild peripheral neuropathy | Low | Weight reduction and improved glycemic control |
| Case 2 (Knee OA + Fall Risk) | Female | 70 | Knee osteoarthritis | Knee pain during walking, prior fall incident | Low | Pain reduction, maintenance of walking ability, and fall prevention |
| Case 3 (Post-Colon CA Recovery) | Male | 60 | Post-colon cancer surgery | Deconditioning, limited walking endurance, fatigue | Low | Physical recovery, fatigue reduction, and improvement of lifestyle habits |
| Case 4 (HTN + T2DM + Obesity) | Female | 68 | Hypertension, type 2 diabetes mellitus, obesity | Dizziness, mild exertional dyspnea, on antihypertensive and antidiabetic medication | Low | Safe improvement of cardiovascular fitness, glycemic control, and gradual weight reduction |
| Case 5 (Fat Loss + Cardiovascular Endurance) | Female | 28 | None (healthy individual) | No clinical limitations; moderate exercise experience | Moderate to High | Body fat reduction and improvement of cardiovascular endurance |
| Case 6 (Muscle Hypertrophy + Strength) | Male | 30 | None (healthy individual) | No clinical limitations; prior resistance training experience | High | Increase muscle mass and improve overall strength |

T2DM = type 2 diabetes mellitus; OA = osteoarthritis; CA = cancer; HTN = hypertension. Cases 1–3 were adopted from Choi et al. [7].

### 2.3. AI Model and Session Control

All outputs were generated automatically through the Vertex AI API (gemini-2.5-flash, Google LLC, Mountain View, CA, USA). Temperature was set to 1.0 to observe output variability arising from the model's stochastic nature under standard generation conditions, and other generation parameters, including top-p, were kept at their default values. Unlike the previous study [7], which involved manual prompt entry through a web interface using a non-logged-in guest session, this study adopted an API-based automated approach to enhance the reproducibility and degree of control over experimental conditions. No additional reasoning elicitation techniques such as chain-of-thought were applied.

### 2.4. Output Generation Procedure

Exercise prescriptions were generated 20 times per scenario under identical prompt conditions, yielding a total of 120 outputs. This repetition count was determined with reference to repeated generation designs used to evaluate LLM response variability [8]. The prompt was based on the guideline-based instruction from the previous study [7], with three modifications: addition of %1RM for resistance exercise intensity notation, removal of the 'ACSM Guidelines for Cancer Survivors' as inapplicable to all scenarios, and removal of the 'if applicable' condition to ensure consistent inclusion of all four exercise components across outputs. The full prompt and case descriptions are provided in Supplementary Material 1. All prompts were kept identical across repetitions.

### 2.5. Data Processing and Preprocessing

All outputs were used in their raw form to reflect the actual response characteristics of the model. No formatting constraints were imposed during the output generation process. The raw outputs contained formatting elements unrelated to the exercise prescription body, including greetings, closing remarks, tables, and bullet points. To prevent such formatting variability from introducing noise into the semantic similarity analysis, a standardized prompt implemented with Claude Sonnet 4.6 (Anthropic API) was applied to consistently extract only the exercise prescription body from all 120 outputs.

### 2.6. Consistency Evaluation

Consistency was evaluated across three dimensions: (1) semantic consistency, (2) structural consistency, and (3) safety consistency.

Semantic consistency was assessed using a pretrained sentence embedding model (all-MiniLM-L6-v2) [12]. Pairwise cosine similarity was computed among the 20 outputs within each scenario (190 pairs), and the mean and standard deviation were used as consistency indicators.

Structural consistency was evaluated based on four FITT components: exercise type, frequency, intensity, and time. Classification was performed using Claude Sonnet 4.6 as an independent LLM evaluator, with the generation and evaluation models separated to mitigate self-evaluation bias [9,11,13]. Intensity was classified for the initial prescription period (weeks 1–4) into three levels (low, moderate, high) using RPE, %HRmax, and MET for aerobic exercise [14,15] and %1RM for resistance exercise [16,17]. Outputs where intensity could not be determined were labeled "estimated" or "unclassifiable," with unclassifiable frequency reported separately (Table 2).

Safety consistency was assessed by evaluating the binary presence (1) or absence (0) of four categories: contraindications, precautions, symptom monitoring, and risk warnings. Contraindications and precautions were based on the safety evaluation rubric of the previous study [7], symptom monitoring referenced the ACSM Guidelines for Exercise Testing and Prescription [18], and risk warnings were operationally defined with reference to the same guidelines. The inclusion rate per scenario served as the primary safety consistency indicator. Additionally, sentence-level counts per category were computed to quantify the extent of safety expression. Full prompts and classification criteria are provided in Supplementary Material 2.

**Table 2.** Intensity Classification Criteria for Aerobic and Resistance Exercise

| Level | Aerobic Exercise | Resistance Exercise |
|---|---|---|
| Low | RPE <12 / %HRmax <64% / <3 METs | <50% 1RM |
| Moderate | RPE 12-13 / %HRmax 64-76% / 3-5.9 METs | 50–69% 1RM |
| High | RPE ≥14 / %HRmax ≥77% / ≥6 METs | ≥70% 1RM |
| Unclassifiable | Expression present but cannot be mapped to above criteria | Expression present but %1RM not stated or estimable |

RPE = rating of perceived exertion (Borg 6–20 scale); HRmax = maximum heart rate; MET = metabolic equivalent of task; 1RM = one-repetition maximum.

*2.7. Quality Verification*

To confirm that preprocessing introduced no semantic distortion, three outputs per scenario (n = 18 total; 15%) were randomly selected and manually reviewed by the first author for preservation of key prescription content, including exercise type, intensity, and safety-related guidelines. Minor formatting differences were observed, but substantive content was confirmed identical before and after preprocessing in all reviewed cases.

*2.8. Statistical Analysis*

Descriptive statistics are presented as mean ± standard deviation. Nonparametric tests were applied given the small sample size per scenario (n = 20). Differences in semantic consistency (cosine similarity) and safety-related sentence counts across scenarios were analyzed using the Kruskal-Wallis test, followed by Dunn's post hoc test with Bonferroni correction when significant differences were identified. Structural consistency was summarized using descriptive statistics of FITT classification frequencies. The significance level was set at $p < 0.05$. All analyses were conducted using Python 3.12.13 (Google Colaboratory, Google LLC) with the following libraries: sentence-transformers, scipy, scikit_posthocs, numpy, pandas, matplotlib, seaborn, anthropic, and **vertexai** libraries for Claude and Gemini API calls, respectively.

## 3. Results

### 3.1 Semantic Consistency Analysis

SBERT (all-MiniLM-L6-v2)-based cosine similarity was computed for 190 pairwise comparisons per scenario, with mean similarity values ranging from 0.879 to 0.939 across all scenarios, indicating generally high consistency (Table 3, Figure 1). The highest consistency was observed in S3 (Mean = 0.939, SD = 0.021) and S2 (Mean = 0.933, SD = 0.022), while S6 showed the lowest consistency and greatest variability (Mean = 0.879, SD = 0.052), suggesting that output consistency increases as clinical constraints become more clearly defined. The Kruskal-Wallis test revealed significant differences across scenarios (H = 328.37, $p < 0.001$), and Dunn's post hoc test showed that S2 and S3 differed significantly from all other scenarios ($p < 0.001$), while no significant differences were observed among S1, S4, S5, and S6 (Figure 2).

**Table 3.** Descriptive Statistics of SBERT-based Cosine Similarity by Scenario

| Scenario | Case Description | N pairs | Mean | SD | Min | Max |
|---|---|---|---|---|---|---|
| S1 | T2DM + Obesity | 190 | 0.887 | 0.041 | 0.78 | 0.974 |
| S2 | Knee OA + Fall Risk | 190 | 0.933 | 0.022 | 0.863 | 0.981 |
| S3 | Post-colon CA Recovery | 190 | 0.939 | 0.021 | 0.877 | 0.988 |
| S4 | HTN + T2DM + Obesity | 190 | 0.895 | 0.036 | 0.771 | 0.966 |
| S5 | Fat Loss + Endurance | 190 | 0.891 | 0.047 | 0.752 | 0.97 |
| S6 | Hypertrophy + Strength | 190 | 0.879 | 0.052 | 0.726 | 0.973 |

T2DM = type 2 diabetes mellitus; OA = osteoarthritis; CA = cancer; HTN = hypertension. Cases 1–3 were adopted from Choi et al. [7].

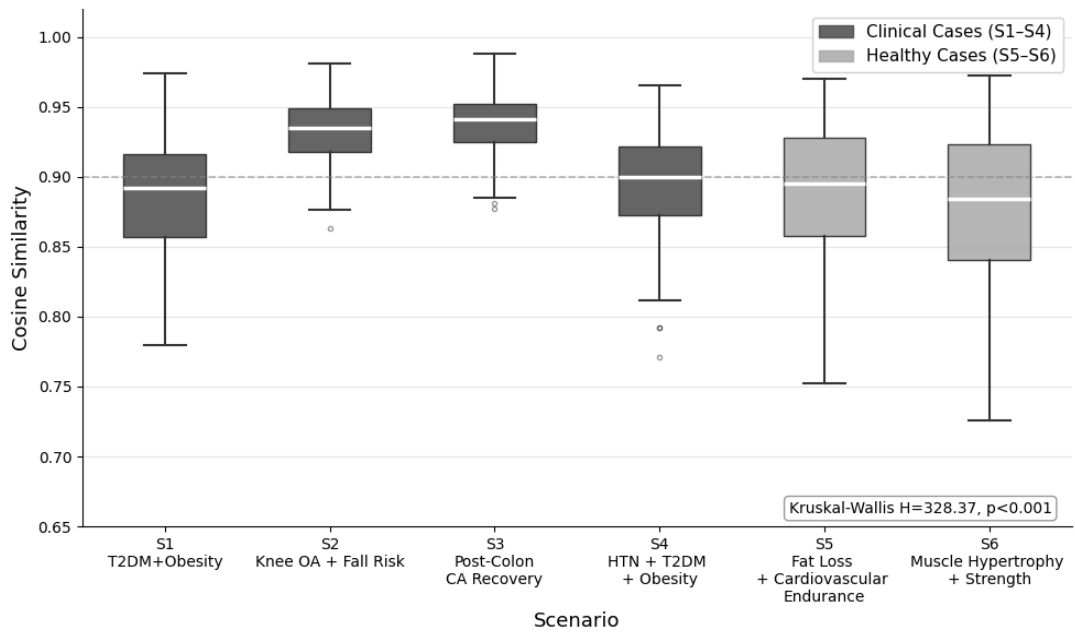

**Figure 1.** Distribution of SBERT-based cosine similarity across scenarios (all-MiniLM-L6-v2). Each box represents 190 pairwise similarity values from 20 repeated outputs. Dark gray = clinical cases (S1–S4); light gray = healthy adult cases (S5–S6). Dashed line indicates 0.90. Kruskal-Wallis H = 328.37, p < 0.001. T2DM = type 2 diabetes mellitus; OA = osteoarthritis; CA = cancer; HTN = hypertension.

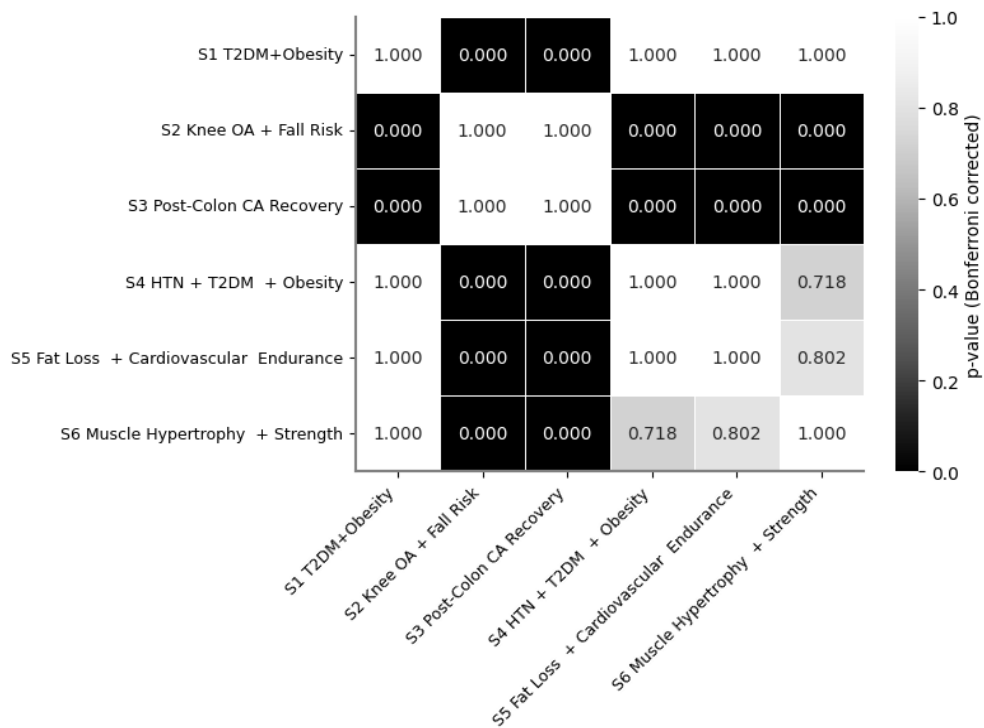

**Figure 2.** Pairwise p-values from Dunn's post hoc test with Bonferroni correction. Black cells indicate significant differences (p < 0.05), gray cells intermediate values, and white cells non-significant comparisons (p ≥ 0.05). T2DM = type 2 diabetes mellitus; OA = osteoarthritis; CA = cancer; HTN = hypertension.

*3.2 Structural Consistency: FITT Classification*

FITT components were classified based on the initial prescription phase (Weeks 1–4) using Claude Sonnet 4.6 as an AI judge (Table 4, Figure 3). In clinical cases (S1–S4), aerobic exercise three to four times per week combined with resistance exercise twice per week was the most common frequency pattern (45–50%), while S5 showed an aerobic-dominant pattern (4–5×/week; 65%) and S6 a resistance-centered pattern (4×/week; 35%). Aerobic intensity in clinical cases was predominantly low (90–100%), whereas healthy adult cases were classified as moderate or high (100%). Resistance exercise intensity was predominantly low in clinical cases (75–85%), with unclassifiable outputs observed in 10–25% of cases. Exercise duration ranged from 10–30 minutes in clinical cases, 30–40 minutes in S5, and 60–90 minutes in S6. Across all scenarios, aerobic, resistance, balance, and flexibility exercises were prescribed in combination, with walking and balance training predominant in clinical cases and resistance training predominant in healthy adult cases.

**Table 4.** FITT Intensity Classification by Scenario (Weeks 1–4)

| Scenario | Frequency | Aerobic Intensity n(%) | Resistance Intensity n(%) | Time |
|---|---|---|---|---|
| S1 | Aerobic 3–4x/Resistance 2x per week (50%) | Low 19(95%), Moderate 1(5%) | Low 15(75%), Moderate 3(15%), Unclass. 2(10%) | 20-30 min |
| S2 | Aerobic 3–4x/Resistance 2x per week (45%) | Low 18(90%), Moderate 2(10%) | Low 17(85%), Unclass. 3(15%) | 10-25 min |
| S3 | Aerobic 3–4x/Resistance 2x per week (50%) | Low 20(100%) | Low 15(75%), Unclass. 4(20%), Moderate 1(5%) | 15-30 min |
| S4 | Aerobic 3–4x/Resistance 2x per week (50%) | Low 20(100%) | Low 15(75%), Unclass. 5(25%) | 10-30 min |
| S5 | Aerobic 4–5x/Resistance 3x per week (65%) | Moderate 20(100%) | Moderate 19(95%), High 1(5%) | 30-40 min |
| S6 | Resistance 4x/Aerobic 2–3x per week (35%) | Moderate 17(85%), High 3(15%) | Moderate 14(70%), High 6(30%) | 60-90 min |

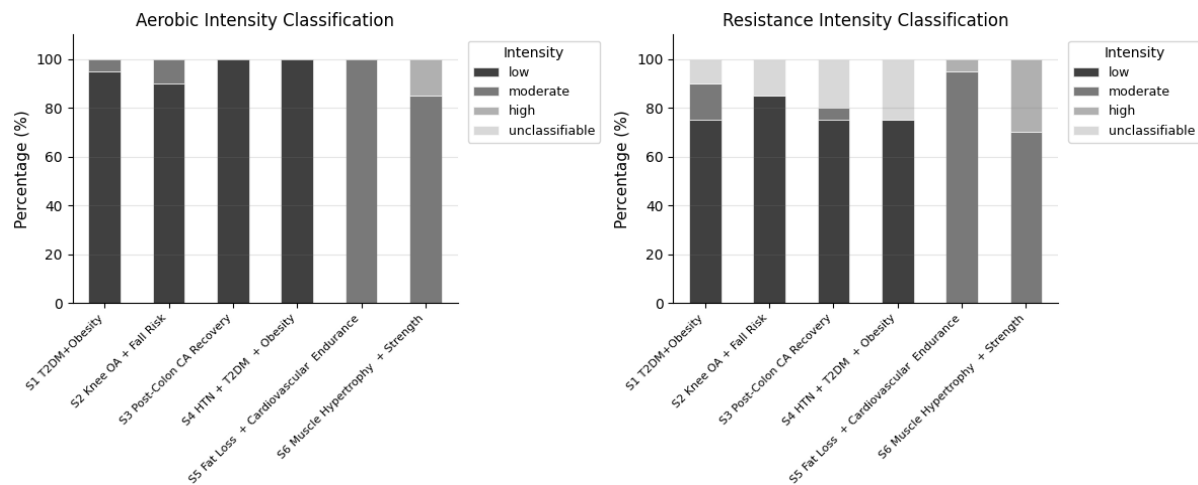

**Figure 3.** Distribution of FITT intensity classifications across scenarios as assessed by Claude Sonnet 4.6 (AI-as-a-Judge).

Left panel shows aerobic intensity; right panel shows resistance intensity. Classifications are based on initial prescription phase (Weeks 1–4). The lightest shade indicates unclassifiable outputs.

*3.3 Safety Expression Consistency*

Evaluation of the presence or absence of safety-related expressions across four categories, including contraindications, precautions, symptom monitoring, and risk warnings, revealed that all four items were included in 100% of the 120 outputs.

Beyond binary inclusion, additional analysis of sentence counts by category revealed significant differences in the total number of safety-related sentences across scenarios (Kruskal-Wallis: H = 86.18, $p < 0.001$). Clinical cases (S1–S4) generated more safety-related expressions than healthy adult cases (S5–S6), with the highest value observed in S4 (Mean = 61.40, SD = 6.62). Significant differences were also identified across all four individual categories (all $p < 0.01$, Table 5).

Dunn's post hoc test with Bonferroni correction revealed that S4 differed significantly from S2, S3, S5, and S6 (all $p < 0.05$). S1 and S3 also showed significantly higher total safety sentence counts compared to S5 and S6 (all $p < 0.05$). No significant differences were observed between S2 and the healthy adult cases (S5, S6), nor between S5 and S6 (Table 5).

**Table 5.** Descriptive Statistics and Kruskal-Wallis Test Results of Safety-Related Sentence Counts by Scenario and Category

| Category | S1 | S2 | S3 | S4 | S5 | S6 | H | p |
|---|---|---|---|---|---|---|---|---|
| Contraindication | 11.90 ± 3.99 | 10.00 ± 3.01 | 14.35 ± 2.82 | 13.95 ± 2.86 | 11.25 ± 6.23 | 10.55 ± 6.01 | 20.55 | 0.001 |
| Precaution | 19.75 ± 4.33 | 15.25 ± 3.74 | 16.70 ± 3.36 | 27.35 ± 3.22 | 10.00 ± 2.13 | 11.90 ± 2.38 | 84.57 | <0.001 |
| Monitoring | 10.70 ± 2.66 | 6.75 ± 1.71 | 6.70 ± 1.46 | 13.50 ± 2.24 | 3.70 ± 1.08 | 3.35 ± 1.35 | 97.76 | <0.001 |
| Warning | 4.35 ± 1.09 | 4.20 ± 1.40 | 4.80 ± 2.29 | 6.60 ± 1.23 | 1.90 ± 0.72 | 2.45 ± 0.76 | 77.11 | <0.001 |
| Total | 46.70 ± 9.95[a] | 36.20 ± 5.63[b] | 42.55 ± 5.55[a] | 61.40 ± 6.62[c] | 26.85 ± 7.35[b] | 28.25 ± 7.30[b] | 86.18 | <0.001 |

Values are presented as mean ± SD (n = 20 per scenario). H = Kruskal-Wallis test statistic. Different superscripts indicate significant differences between scenarios based on Dunn's post hoc test with Bonferroni correction (p < 0.05).

## 4. Discussion

This study systematically evaluated the repeated generation consistency of LLM-based exercise prescriptions produced by Gemini 2.5 Flash across three dimensions: semantic consistency, FITT structural classification, and safety expression consistency. The results confirmed that overall semantic consistency was high, while variability was observed in certain structural components, suggesting that the reliability of LLM outputs is influenced not only by model characteristics but also by prompt structure and imposed constraints. This finding has direct clinical implications: the same patient profile may yield structurally different prescriptions across repeated generations, potentially resulting in non-equivalent exercise stimuli that could affect safety and treatment outcomes in real-world deployment.

Research on repeated generation consistency of LLMs has been conducted primarily in clinical decision support and medical question answering. Carandang et al. [19] reported high semantic consistency in repeatedly generated clinical notes, aligning with the cosine similarity values observed in the present study (0.879–0.939), and Hanss et al. [20] similarly found high consistency in GPT-4 responses, suggesting that response consistency may serve as a useful reliability indicator. In the exercise prescription domain, Lai et al. [5] called for standardized evaluation frameworks to ensure consistency and reliability across AI models. The present study directly addresses this gap by proposing an evaluation pipeline combining SBERT-based semantic similarity with AI-as-a-Judge FITT classification, extending the methodological scope of LLM-based exercise prescription research by introducing repeated generation consistency as a reliability dimension.

Consistency was higher in scenarios with more clearly defined clinical constraints. The highest consistency was observed in S3 (Mean = 0.939) and S2 (Mean = 0.933), whereas S6 showed the lowest consistency (Mean = 0.879) and greatest variability (SD = 0.052), suggesting that LLM outputs tend to converge under well-defined clinical conditions while output diversity increases when the range of clinically valid options is broader. S1 and S4, despite being clinical cases, showed relatively lower consistency than S2 and S3, likely reflecting the broader prescription latitude associated with metabolic and multimorbidity conditions. Notably, S2 did not differ significantly from healthy adult cases in the post hoc analysis, suggesting that consistency is shaped more by the degree of guideline constraint than by clinical status alone. Kruskal-Wallis and Dunn post hoc analyses confirmed that S2 and S3 differed significantly from all other scenarios (p < 0.001), indicating that case type should be considered as a control variable in future research.

It is also important to note that high semantic similarity does not necessarily imply clinical equivalence. SBERT-based similarity reflects overall semantic structure but may not capture clinically meaningful numerical variations, such as a shift in aerobic intensity from 50% to 70% HRmax. Semantic consistency should therefore be complemented by quantitative evaluation approaches such as the FITT classification applied in this study.

The FITT classification results indicate that Gemini 2.5 Flash consistently reflected international exercise prescription guidelines according to case characteristics. Clinical cases (S1–S4) were predominantly prescribed low-intensity aerobic (90–100%) and resistance exercise (75–85%), while a clear shift toward moderate-to-high intensity was observed in healthy adult cases (S5–S6), consistent with ACSM recommendations [18].

However, variability was observed in several quantitative components. Resistance exercise intensity showed the greatest instability, with unclassifiable outputs in 10–25% of clinical cases despite explicit prompting for %1RM values, and inconsistent classifications within the same scenario in S6. Frequency variability was also notable in S6, where the most common pattern accounted for only 35% of outputs. Session duration was not consistently stated, with some outputs providing estimated rather than explicit values. These patterns of quantitative instability are consistent with findings by Qiu et al. [21] on LLM performance decline in treatment planning tasks, and suggest that both prompt structure and inherent model behavior contribute to output variability, underscoring the need for structured prompt design alongside continued evaluation of model-level characteristics. Future studies comparing multiple LLMs under identical prompt conditions would help clarify whether these patterns are model-specific or reflect broader characteristics of generative AI in exercise prescription.

The 100% inclusion rate of safety-related expressions reflects the effectiveness of explicit safety instructions in the prompt, suggesting that structured prompt design is effective in ensuring safety content. However, the quantity of safety expressions varied significantly by scenario (H = 86.18, p < 0.001), with S4 generating the highest volume (Mean = 61.40), and precaution and monitoring showing the greatest variability. Dunn's post hoc analysis revealed that S2 did not differ significantly from healthy adult cases, indicating that safety expression volume is determined more by clinical complexity than by clinical status alone. As emphasized by previous study [22] and further supported by evidence from the exercise prescription domain [7], expert validation remains essential prior to clinical deployment.

Although the LLM-as-a-Judge paradigm is emerging as a scalable and cost-effective assessment alternative, careful design and standardization are required to ensure reliability [23]. The present study constitutes an early systematic attempt to apply this methodology within the exercise prescription domain. Several limitations should be acknowledged. First, only a single model was evaluated; future studies should incorporate multi-model comparisons. Second, safety evaluation was limited to the presence and quantity of safety-related expressions, and output consistency does not imply clinical validity. Third, the reliability of LLM-based classification was not formally validated against expert judgment. Fourth, the six scenarios may not fully represent the diversity of clinical practice.

## 5. Conclusion

This study evaluated the repeated generation consistency of LLM-generated exercise prescriptions across three dimensions: semantic similarity, FITT structural classification, and safety expression consistency. High semantic consistency was observed across all scenarios, with greater consistency in cases with more clearly defined clinical constraints. However, variability in key quantitative components, particularly exercise intensity, suggests inherent limitations in generating stable numerical prescriptions. These findings indicate that the reliability of LLM-based exercise prescription depends substantially on prompt structure and imposed constraints, and that clinical deployment requires structured decision-making processes and quantitative control mechanisms in addition to generative models.

**Declaration of generative AI and AI-assisted technologies in the manuscript preparation process**
During the preparation of this work, the authors used AI-based language tools to assist with translation and language refinement. After using these tools, the authors reviewed and edited the content as needed and take full responsibility for the content of the published article.

**Funding**
This research received no external funding.

**Conflict of Interest**
The author declares no conflict of interest.